%% file: main.tex
\documentclass{article} % For LaTeX2e
\usepackage{collas2022_conference,times}
\usepackage{multirow}

\input{math_commands.tex}

\usepackage{hyperref}
\hypersetup{
    colorlinks=true,
    linkcolor=red,
    filecolor=magenta,      
    urlcolor=blue,
    citecolor=purple,
    pdftitle={Overleaf Example},
    pdfpagemode=FullScreen,
    }

\usepackage{graphicx}
\usepackage{caption}
\usepackage{comment}
\newcommand{\Scd}[1]{\textit{{\color{blue} #1}}}

\title{optimal transport meets noisy label robust loss and MixUp regularization for domain adaptation}

\author{Kilian Fatras  \thanks{corresponding author: kilian.fatras@mila.quebec} \\
Mila - Quebec AI Institute, McGill University\\
Montr\'eal, Qu\'ebec, Canada \\
\And % Use And to have authors side by side
Hiroki Naganuma \& Ioannis Mitliagkas  \\
Mila - Quebec AI Institute, Universit\'e de Montr\'eal\\
Montr\'eal, Qu\'ebec, Canada \\
}

 \collasfinalcopy % Uncomment for camera-ready version, but NOT for submission.

\begin{document}

\maketitle

\begin{abstract}
It is common in computer vision to be confronted with domain shift: images which have the same class but different acquisition conditions.
In domain adaptation (DA), one wants to classify unlabeled target images using source labeled images.
Unfortunately, deep neural networks trained on a source training set perform poorly on target images which do not belong to the training domain. One strategy to improve these performances is to align the source and target image distributions in an embedded space using optimal transport (OT). However OT can cause negative transfer, \emph{i.e.} aligning samples with different labels, which leads to overfitting especially in the presence of label shift between domains.
%In this work, to mitigate these negative alignments, we propose seeing them as assigning noisy labels to target images and regularizing the neural network to mitigate their effect.
In this work, we mitigate negative alignment by explaining it as a noisy label assignment to target images.
We then mitigate its effect by appropriate regularization.
We propose to couple the MixUp regularization \citep{zhang2018mixup} with a loss that is robust to noisy labels in order to improve domain adaptation performance. We show in an extensive ablation study that a combination of the two techniques is critical to achieve improved performance.
Finally, we evaluate our method, called \textsc{mixunbot}, on several benchmarks and real-world DA problems.
\end{abstract}

\section{Introduction}

Deep neural networks have reached state-of-the-art performance on classification problems due to their ability to fit complex dataset while also generalizing within a specific domain \citep{He2016}. However, in applications like computer vision, it is standard to have same-class samples coming from different domains, for instance when they have different backgrounds or colorspaces. Unfortunately, the generalization of deep neural networks across different domains is poor, and the object of intense research.
Diversifying the domains with new samples in the training dataset is also a challenging task. For instance in medicine, collecting and annotating data is time-consuming and prone to errors.
The problem of \emph{domain adaptation}, tackles the case where we have access to two domains sharing the same classes where one has labeled data, called the source domain, and the other has unlabeled data, called the target domain. The purpose of domain adaptation problems is to classify the unlabeled target samples using the labeled source samples \citep{Pan2010, Patel2015}. A popular and efficient method to solve this problem is to use the \emph{alignment strategy}, where same-class samples from different domains are aligned in an embedded space. To align the embedded source and target samples, several techniques exist such as adversarial training \citep{DANN, cdan2018, alda2020} or optimal transport  \citep{Courty_OTDA, courty_jdot, redko2017}.

Optimal Transport \citep{COT_Peyre} has become one of the most used methods to compare probability distributions in machine learning. It has been popular for tasks such as generative models \citep{arjovsky17a, genevay18a, salimans2018improving, burnel2021} or supervised learning problems \citep{frogner_2015, Fatras2021WAR}. In domain adaptation, it has been used to transport the source domain to the target domains \citep{Courty_OTDA}, or to align the domains in a joint space of data and labels \citep{courty_jdot, Damodaran_2018_ECCV}. To reduce the OT cost, \cite{Damodaran_2018_ECCV} used a minibatch approximation of OT \citep{fatras20a}, which led to non-optimal connections between domains due to the minibatch samplings and the marginal constraints of exact OT. 
This phenomenon corresponds to negative transfer in domain adaptation problems. To mitigate the non-optimal connections of minibatch OT, \textsc{deepjdot} \citep{Damodaran_2018_ECCV} required large batch sizes---otherwise small batch sizes lead deep neural networks to overfit. Another workaround to mitigate the non-optimal connections was to use Unbalanced Optimal Transport (UOT), an OT variant with relaxed marginals, as proposed in \cite{fatras21a}.

In this paper, we explain non-optimal connections  as assigning noisy labels to target samples. To avoid overfitting from noisy labels, we propose to regularize the neural networks and to use a noisy label robust loss in the transfer term. We propose to couple two techniques: \emph{i)} regularize the neural networks using the MixUp regularization \citep{zhang2018mixup}  on both source and target domains, MixUp interpolates samples from the same distribution uniformly at random as well as their label when available; \emph{ii)} Use the symmetric cross-entropy loss in the transfer term, which has been proven to be robust to label noise \citep{wang2019symmetric}. Our findings show that it is the combination of the MixUp and symmetric cross-entropy which leads to an increase in the performances of models as shown in an extensive ablation study, while the use of the techniques separately does not lead to any increase. It can be used for different optimal transport loss as we show in our experiments.

%In this paper, we explain non-optimal connections  as assigning noisy labels to target samples. In order to mitigate the influence of these noisy labels and to avoid overfitting, we propose to regularize the neural networks using MixUp \citep{zhang2018mixup} on both source and target domains with a robust to noisy label loss function between the target prediction and the source label. MixUp regularization interpolates samples from the same distribution uniformly at random as well as their label. Our findings show that it is the combination of the two approaches which leads to an increase in the performances of models as shown in an extensive ablation study, while the use of the techniques separately does not lead to any increase. It can be used for different optimal transport loss as we show in our experiments.

This paper is structured as follows. In Section \ref{sec:background}, we review the different methods to solve domain adaptation and we define optimal transport as well as its use in domain adaptation problems. In Section \ref{sec:proposed_mixot}, we present our method \textsc{mixunbot} as well as the components it is built upon such as \textsc{deepjdot}, MixUp, and the symmetric cross-entropy (SCE) loss. In Section \ref{sec:exp}, we present extensive domain adaptation and partial domain adaptation experiments.

\section{Related work on domain adaptation and optimal transport}\label{sec:background}

In this section, we review the basic strategies for solving domain adaptation problems. Then we define formally optimal transport and we discuss its empirical implementation in deep learning applications, its current limitations and its use in domain adaptation. 

\paragraph{Unsupervised domain adaptation and notations}
We recall that in unsupervised domain adaptation, we want to classify unlabeled target data using labeled source data \citep{Pan2010}. Formally, let $\mathcal{D}_s$ (resp. $\mathcal{D}_t$) be the labeled source (resp.\ target) dataset which is composed of $n$ \emph{i.i.d} random labeled (resp.\ unlabeled) vectors in $\R^d$ drawn from a distribution $ \mu $ (resp. $\nu$), \emph{i.e.,} $\mathcal{D}_s = \big\{ (\xx_i^s, \yy_i^s) \big\}_{i=1}^n, \xx_i^s \in \mathbb{R}^d$ (resp. $\mathcal{D}_t = \big\{ (\xx_j^t) \big\}_{j=1}^n, \xx_j^t \in \mathbb{R}^d$). We denote the tuple of source (resp.\ target) vectors as $\boldsymbol{X}=(\xx_1, \cdots, \xx_n) $ (resp.\ $\boldsymbol{Z}=(\zz_1, \cdots, \zz_n) $) and the product measure $\mu^{\otimes m }$ denotes a sample of $m$ independent random variables following the distribution $\mu$. When a map $g$ is applied to a tuple $\XX$, we have $g(\XX)=\{g(\xx_1), \cdots, g(\xx_n)\}$. In vanilla domain adaptation, both domains share the same label space $\mathcal{Y}_s=\mathcal{Y}_t$. Other variants can be considered such as partial domain adaptation \citep{Cao_2018_ECCV}, where some source classes are not in the target domain $\mathcal{Y}_t \subset \mathcal{Y}_s$ and open set domain adaptation \citep{Busto2017}, where some target classes are not in the source domain $\mathcal{Y}_s \subset \mathcal{Y}_t$. In this work, we focus on vanilla and partial domain adaptation.

Several strategies exist to classify the target samples using deep neural networks, and they can be divided into three groups.
The first group introduces a domain discriminator, which predicts if a sample is from the source or the target domain, and it is trained in an adversarial training manner as invented in \cite{goodfellow_gan}. The adversarial training encourages domain confusion in order to fool the discriminator so that it can not decide which domain the samples come from \cite{DANN}. 
Several variants of the initial idea were published, for instance \cite{cdan2018, alda2020}.
Another group of methods is the reconstruction of both source and target samples. In \cite{Ghifary2016}, authors learn an embedding where source samples are classified, and target samples are encoded, then the learned representation not only preserves the ability to discriminate but also encodes useful information from the target domain. The final group of methods aims at minimizing the domain discrepancy by minimizing different distances between probability distributions like maximum mean discrepancy (MMD) \citep{long15, long17a} or optimal transport \citep{Courty_OTDA, courty_jdot, Damodaran_2018_ECCV, fatras21a}. Our methods are built upon the works on optimal transport, and its definition is the topic of the next paragraph.

\paragraph{Optimal transport}
The optimal transport cost measures the minimal displacement cost of moving a probability distribution $\mu$ to another probability distribution $\nu$ with respect to a ground metric $c$ on the data space $\mathcal{X}$ and $\mathcal{Z}$ \citep{COT_Peyre}. Formally, let $(\mu, \nu) \in \mathcal{M}_{+}^1(\mathcal{X}) \times \mathcal{M}_{+}^1(\mathcal{Z})$, where $\mathcal{M}_{+}^1(\mathcal{X})$ (resp.\ $\mathcal{M}_{+}^1(\mathcal{Z})$) is the set of probability measures on $\mathcal{X}$ (resp. $\mathcal{Z}$). For a ground cost $c:\mathcal{X}\times \mathcal{Z} \mapsto \mathbb{R}_+$, the optimal transport cost between the two distributions $\mu, \nu$ is 
\begin{equation}
    W_{c}(\mu, \nu) = \underset{\pi \in \mathcal{U}(\mu, \nu)}{\text{min}} \int_{\mathcal{X}\times \mathcal{Z}}c(\xx,\zz) d\pi(\xx,\zz),
\label{eq:wasserstein_dist}
\end{equation}
where $\mathcal{U}(\mu, \nu)$ is the set of joint probability distributions with
marginals $\mu$ and $\nu$ such that 
$
\mathcal{U}(\mu, \nu) = \left \{ \pi \in \mathcal{M}_{+}^1(\mathcal{X}, \mathcal{Z}): \PP_{\mathcal{X}}\#\pi = \mu, \PP_{\mathcal{Z}}\#\pi = \nu \right\}\nonumber
$. {$\PP_{\mathcal{X}}\#\pi$ (resp.\ $\PP_{\mathcal{Z}}\#\pi$) is the marginalization of $\pi$ over $\mathcal{X}$ (resp.\ $\mathcal{Z}$)}. These constraints enforce that all the mass from $\mu$ is transported to $\nu$ and vice-versa. This optimization problem is called the Kantorovitch formulation of OT and the optimal $\pi$ is called an optimal transport plan.  The ground cost $c$ is usually chosen as the Euclidean or squared Euclidean distance on $\R^d$. In the case where $c$ is a distance, $W_{c}$ is a distance between probability distributions called the Wasserstein distance. When the distributions are discrete, the problem becomes a discrete linear program that can be solved with a cubical complexity in the size of the distributions support, which can be prohibitive for applications with large datasets.

That is why the optimal transport cost has been computed in practice between random minibatches of input distributions \citep{genevay18a, Damodaran_2018_ECCV}. The minibatch approximation of optimal transport has been theoretically investigated as a loss function in \citep{fatras20a, fatras2021minibatch} and its deviation from exact OT in \citep{mbot_Sommerfeld}. Instead of solving the optimal transport cost between the full input distributions, the minibatch approximation computes the expectation of optimal transport cost between minibatches drawn uniformly at random. While it defines a transport problem in the sense that all the mass of input distributions are transported, \cite{fatras20a} nonetheless showed that minibatch OT acts like a regularization on the transportation plan, leading to a denser plan. Those are due to the non-optimal connections from the random sampling of minibatches and the marginal constraints in the OT definition. They also showed that the number of connections increases when the minibatch size decreases. The consequences are that minibatch OT is not a distance but has some appealing statistical and optimization properties. Despite these weaknesses, optimal transport and its minibatch approximation have found numerous applications to compare probability distributions from different domains as in domain adaptation. %The ability of optimal transport to compare probability distributions justified its use in domain adaptation to compare the source and target domains. 

\paragraph{Optimal transport for domain adaptation}
In domain adaptation, optimal transport has been used to transport the source domain to the target domain with theoretical guarantees \citep{redko2017}. In \cite{Courty_OTDA}, source samples were transported with their labels to the target domain and a classifier was then trained on the transported labeled samples. It was achieved by using the optimal transport plan $\pi$ as a barycentric mapping between the source and the target distributions. A more recent strategy was to align the joint distributions of features and labels of the different domains \citep{courty_jdot, Damodaran_2018_ECCV} instead of only considering the feature distributions. To learn the joint space, OT based techniques used a ground cost which incorporates a term on the (embedded) sample space as well as a term on the label space. This OT cost enables one to learn a mapping which tries to assign labels to samples which have close embeddings. However, OT can make connections between samples from different classes notably in the case of label shift, when there is a different proportion of labels between domains \citep{redko19a}. This problem is exacerbated when one uses the minibatch OT approximation to reduce the OT complexity as done in \cite{Damodaran_2018_ECCV}, where the non-optimal connections between distributions from the minibatch approximation led to even more negative transfer in domain adaptation problems. Indeed by sampling samples uniformly at random, it is likely to draw minibatches which have a shift in the labels especially when there is a shift in the labels between the full distributions. For instance, drawing a source minibatch full of images from one class and a target minibatch full of samples from another class. Thus, the \textsc{deepjdot} method requires large batch sizes in order to be performing and this is inconvenient as the purpose of using minibatches is to decrease the OT cubical complexity. In order to mitigate the effect of these connections, \citep{fatras21a} proposed to rely on unbalanced optimal transport at the minibatch level that we formally define. Consider $\KL$ to be the Kullback-Leibler divergence, we define the unbalanced OT cost between two probability distributions $\mu, \nu$ as:
\begin{equation}
    \operatorname{UOT}_{c}^\tau(\mu, \nu) = \underset{\pi \geq 0}{\text{min}} \int_{\mathcal{X}\times \mathcal{Z}}c(\xx,\zz) d\pi(\xx,\zz) + \tau(\KL(\pi|\mu) + \KL(\pi|\nu)).
\label{eq:uot_cost}
\end{equation}
It is an OT variant with relaxed marginal constraints which does not transport the mass when it is costly to transport, at the cost of choosing well one extra hyper-parameter, thus decreasing the non-optimal connections from the shift in the labels between domains and the minibatch sampling. However, it does not avoid all non-optimal connections. In the next section, we present our method which aims at regularizing the neural networks in order to mitigate the effect of the non-optimal connections.

\section{Proposed method}\label{sec:proposed_mixot}
We are now ready to present our proposed approach to solve the domain adaptation problem. We start by introducing the different methods which are at the heart of our approach.
\subsection{Background}

\paragraph{\textsc{deepjdot}} Our approach is built upon \textsc{deepjdot} \citep{Damodaran_2018_ECCV} and \textsc{jumbot} \citep{fatras21a}. As described in the previous section, \textsc{deepjdot} aims at finding a joint distribution map between the source and the target distributions. It does so thanks to the ground cost used in the optimal transport problem which incorporates a term on a neural network embedding space and a term on the label space between domains. Formally, let
$g_\theta$  be a feature extractor where the input is mapped from the space of images to a latent
space and $f_\phi$ which maps the latent space to the label space. The embedding space in our experiment is the penultimate layer of a neural network. For two probability distributions $\mu, \nu$, feature extractor $g_\theta$ and classification
map $f_\phi$, the transfer term is:
\begin{align}\label{def:jdot}
  &\mathbb{E}_{(\XX^s, \YY^s) \sim \mu^{\otimes m}, \XX^t \sim \nu^{\otimes m}}h_{C_{\theta, \phi}}((g_\theta(\XX^s),\YY^s),(g_\theta(\XX^t),f_\phi(g_\theta(\XX^t)))), \\
  &\text{with } (C_{\theta, \phi})_{i,j}=\eta_1 \|g_\theta(\xx_i^s) - g_\theta(\xx_j^t)\|_2^2 + \eta_2 \mathcal{L}(\yy_i^s, f_\phi(g_\theta(\xx_j^t))) \nonumber,
\end{align}
where $h$ is the considered OT variant, $\mathcal{L}(.,.)$ is the cross-entropy loss and $\eta_1, \eta_2$ are positive constants. Note that we make an abuse of notation between a tuple of samples $\XX$ and its empirical distribution. We also add a cross-entropy term on the source data. Hence our optimization problem is:
\begin{align}
   \underset{\theta, \phi}{\operatorname{min}} &\sum_i \mathcal{L}(f_\phi(g_\theta(\xx_i^s)), \yy_i^s) + \eta_3 \mathbb{E}_{(\XX^s, \YY^s) \sim \mu^{\otimes m}, \XX^t \sim \nu^{\otimes m}} h_{C_{\theta, \phi}}((\XX^s,\YY^s),(\XX^t,f_\phi(g_\theta(\XX^t)))). %\langle \Pi^\star, C \rangle \Big ),%(g_\theta(\xx_i), g_\theta(\xx_j), \yy_i, f_\phi(g_\theta(\xx_j))
\end{align}
%With $C$ defined as in \eqref{def:jdot}. 
 Intuitively, the OT plan would send mass between source and target samples which have close embeddings as well as close classifications. Thus close embeddings should have the same classification and samples which have the same classification should have close embeddings. However, due to the hard marginal constraints of exact OT, the shift in the labels which occurs between the source and the target domains leads to transport samples from the source to the target samples which have a different label. This is especially the case in partial domain adaptation where there is a serious shift in the labels (classes which are in the source domain but not in the target domain). Furthermore, the minibatch approximation of OT strengthens the non-optimal connections because of the sampling effect.
To address this issue, \cite{fatras21a} developed \textsc{jumbot} which relied on the use of unbalanced OT, which mitigates the transfer of mass when a sample is costly to transport. The only difference between the methods \textsc{jumbot} and \textsc{deepjdot} is the use of entropically regularized unbalanced OT instead of exact OT. Another practical advantage of unbalanced OT is that it can be used to solve the partial domain adaptation problem \citep{fatras21a}.

\begin{figure}[t]
    \centering
        \includegraphics[width=1.\linewidth]{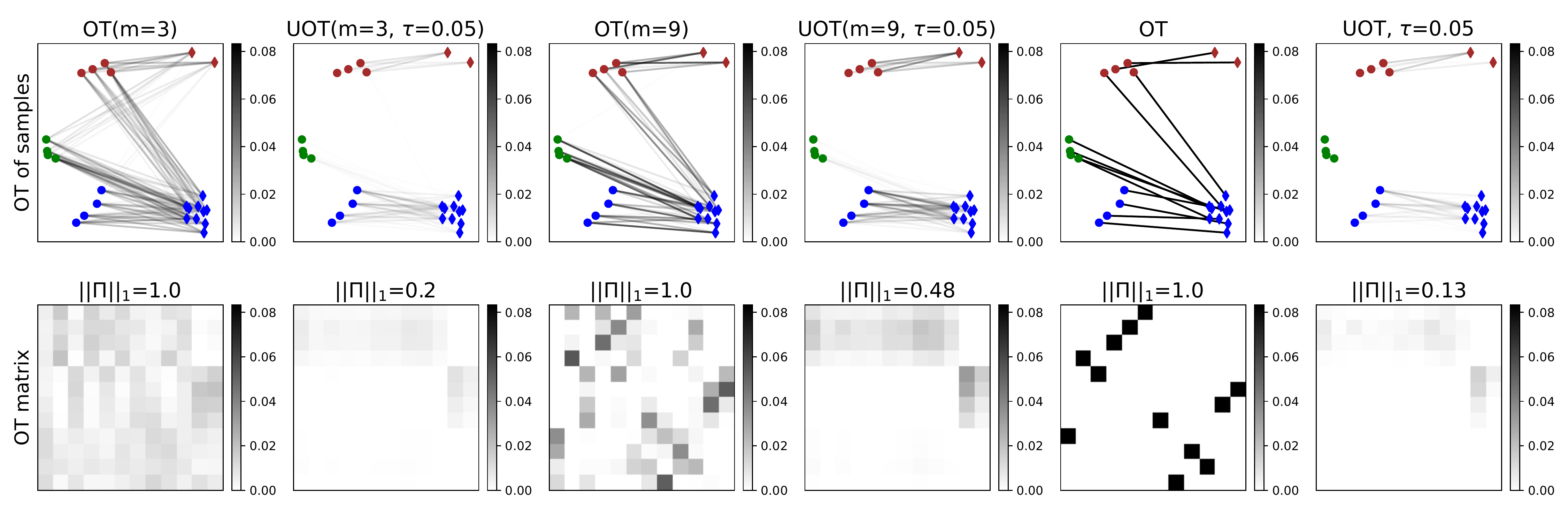}
    %\vspace{-0.5cm}
    \caption{Overview of OT plans between 2D distributions with 12 samples. The distributions have a cluster-like structure, where the source distribution has three (3) balanced clusters and the target distribution has only two (2) imbalanced clusters which are in the source domain. The first row shows the transported mass as connections between samples for different batch size, the second row illustrates the minibatch OT and UOT plans $\Pi$ as well as their mass.}
    \label{fig:mbot_connections}
    %\vspace{-0.5cm}
\end{figure}

Unfortunately, as discussed in \cite{fatras21a}, there is still a transport of mass between samples of different labels. To illustrate this phenomenon, in Figure \ref{fig:mbot_connections} we consider the partial domain adaptation problem in a 2D scenario with 12 samples. The source distribution has three classes and the target distribution has only two classes which are in the source domain. The classes have a cluster-like structure and are balanced in the source domain and imbalanced in the target domain (10 elements for the blue class and only 2 for the brown class). The OT cost and minibatch OT cost create a lot of connections between samples from different classes due to marginal constraints and minibatch sampling for the latter. However, the unbalanced OT based variants suffer less from these connections at the price of choosing a regularization coefficient $\tau$. Furthermore, for small $\tau$ values, the mass of the transport plan can be small as shown in Figure \ref{fig:mbot_connections}, thus decreasing the transfer between domains. To mitigate this problem, \cite{fatras21a} used an extra parameter before the OT term to increase the strength of the transfer loss. To help the transfer between domains, we propose to regularize neural networks to mitigate the influence of non-optimal connections. 
\paragraph{MixUp regularization}
is a training strategy based on a data augmentation technique proposed in \cite{zhang2018mixup}. It artificially augments the training dataset by interpolating samples randomly between them, as well as their labels when they are available, in order to create new training samples. As we work on a joint distribution, our new neighbourhing source distribution considers the interpolated samples and their label while the neighbourhing target distribution considers the interpolated sample and their prediction from the neural network. They are defined as follows:
%\begin{align}
%  \tilde{\mu}\left(\tilde{x}^s, \tilde{y}^s | x_{i}^s, y_{i}^s\right)=\frac{1}{n} \sum_{j}^{n} \mathbb{E}_{\lambda \sim \beta(\alpha)}\left[\delta\left(\tilde{x}^s=\lambda \cdot x_{i}^s+(1-\lambda) \cdot x_{j}^s, \tilde{y}^s=\lambda \cdot y_{i}^s+(1-\lambda) \cdot y_{j}^s\right)\right] \nonumber\\
%  \tilde{\nu}\left(\tilde{x}^t, \tilde{y}^t | x_{i}^t, y_{i}^t\right)=\frac{1}{n} \sum_{j}^{n} \mathbb{E}_{\lambda \sim \beta(\alpha)}\left[\delta\left(\tilde{x}^t=\lambda \cdot x_{i}^t+(1-\lambda) \cdot x_{j}^t, \tilde{y}^t=\lambda \cdot f(x_{i}^t)+(1-\lambda) \cdot f(x_{j}^t)\right)\right]
%\end{align}
\begin{align}
  \tilde{\mu}=\frac{1}{n} \sum_{i,j=0}^{n} \mathbb{E}_{\lambda \sim \beta(\alpha)}\delta_{(\lambda \cdot x_{i}^s+(1-\lambda) \cdot x_{j}^s, \lambda \cdot y_{i}^s+(1-\lambda) \cdot y_{j}^s)} = \mathbb{E}_{\lambda \sim \beta(\alpha)} \tilde{\mu}_\lambda, 
\end{align}
\begin{align}
  \tilde{\nu}=\frac{1}{n} \sum_{i,j=0}^{n} \mathbb{E}_{\lambda \sim \beta(\alpha)}\delta_{(\lambda \cdot x_{i}^t+(1-\lambda) \cdot x_{j}^t, f(\lambda \cdot x_{i}^t+(1-\lambda) \cdot x_{j}^t))} = \mathbb{E}_{\lambda \sim \beta(\alpha)} \tilde{\nu}_\lambda. 
\end{align}
Where $\tilde{\mu}_\lambda$ (resp. $\tilde{\nu}_\lambda$) is the measure associated to interpolated samples from $\mu$ (resp. $\nu$) for a fixed $\lambda$ and $n$ the number of samples in measures $\mu$ and $\nu$. The parameter $\lambda$ follows a beta distribution of parameter alpha $\beta(\alpha)$, where $\alpha$ is set to 0.2 in our experiment. We study its impact on the performance in a sensitivity analysis (see Section \ref{par:analysis}). While it is a simple technique, it has allowed neural networks to reach state-of-the-art results on classification tasks while being more robust to adversarial examples \citep{zhang2021how}. Moreover, MixUp  regularization acts like four different regularizations: perturbed data, temperature scaling, label smoothing, and Jacobian regularization \citep{Carratino2020}. MixUp also finds adaptively the good parameters to make the four regularization work together.

MixUp has been previously used in domain adaptation \citep{mao2019virtual}, however authors coupled it with the method DIRT-T \citep{shu2018dirtt} and the case where it was coupled with optimal transport has not been studied. %The MixUp startegy is applied to the source domain where both samples and labels are interpolated and in the target domain, MixUp is applied on samples only as we do not have access to the labels. Thus we have new distributions $\tilde{\mu}, \tilde{\nu}$. 
MixUp data augmentation was also applied between samples from different domains in \cite{xu2020adversarial}. It was also used in variants of domain adaptation as semi-supervised DA \citep{yang2021deep}.

In this work, we propose to compute the OT cost between the neighbourhing probability distributions $\tilde{\mu}$ and $\tilde{\nu}$ to inherit the benefits from the MixUp regularization on both domains. However at each iteration, we compute the OT cost between minibatches from $\tilde{\mu}_\lambda$ and $\tilde{\nu}_\lambda$ for a fixed $\lambda \sim \beta(\alpha)$, as done in \cite{zhang2018mixup}. While it works well in practice, this implementation leads us to compute an upper bound of $W(\tilde{\mu}, \tilde{\nu})$ as shown in the following Proposition:

\begin{prop}\label{prop:upperbound}
Let $\mu,\nu$ be two probability distributions and $\tilde{\mu}, \tilde{\nu}$ their neighbourhing respective distributions. We have the following upper bound for the Wasserstein distance: 
\begin{equation}
    W(\tilde{\mu}, \tilde{\nu}) \leq \mathbb{E}_{\lambda, \lambda^{\prime}} W(\tilde{\mu}_\lambda, \tilde{\nu}_{\lambda^\prime}).
\end{equation}
\end{prop}

%\begin{align}
 % \mu_1 =\frac{1}{n} \sum_{j}^{n}\left[\delta\left(\tilde{x}^s=\lambda \cdot x_{i}^s+(1-\lambda) \cdot x_{j}^s, \tilde{y}^s=\lambda \cdot y_{i}^s+(1-\lambda) \cdot y_{j}^s\right)\right] \\
  %\text{and } \nu_1 =\frac{1}{n} \sum_{j}^{n}\left[\delta\left(\tilde{x}^t=\lambda \cdot x_{i}^t+(1-\lambda) \cdot x_{j}^t, \tilde{y}^t=\lambda \cdot f(x_{i}^t)+(1-\lambda) \cdot f(x_{j}^t)\right)\right]
%\end{align}
\begin{figure}[t]
    \centering
        \includegraphics[width=0.9\linewidth]{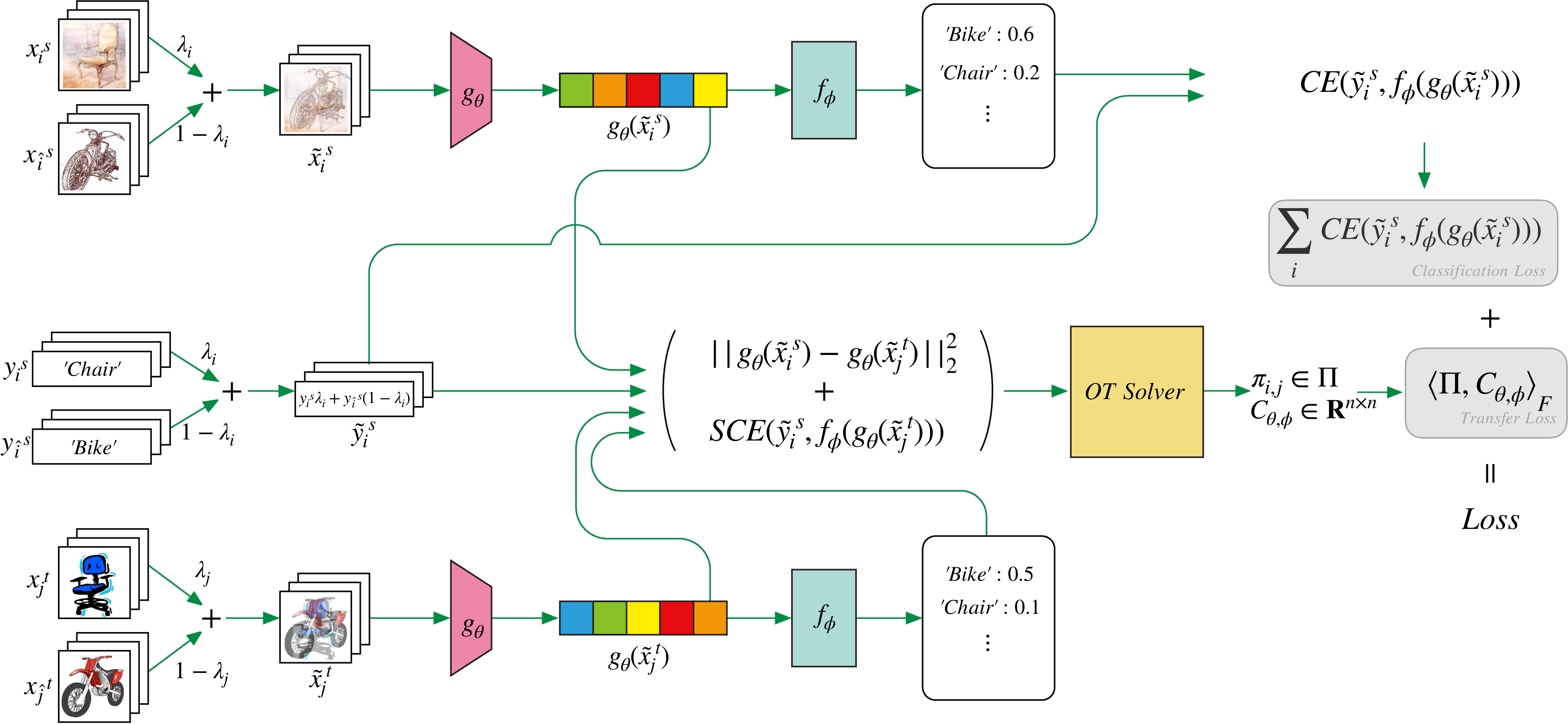}
    \caption{(Best viewed in color) Overview of the proposed \textsc{mixunbot} method between full source and target distributions (we use a minibatch computation in practice). We denote the feature extractor $g_{\theta}$ and the classifier $f_{\phi}$. The latent representations and labels are used to compute a ground cost matrix that is fed to an optimal transport solver. We then minimize a cross-entropy term on the source domain and an optimal transport alignment term between domains.}
    \label{fig:mixot}
\end{figure}

The proof is straightforward thanks to the convexity of the Wasserstein distance \citep[Section 9.1]{COT_Peyre} and Jensen inequality. In this paper, we empirically show that adding MixUp to \textsc{deepjdot} only improves slightly the performances of \textsc{deepjdot} contrary to the findings from \cite{mao2019virtual}, where MixUp really increased the results of DIRT-T. While the use of the MixUp data augmentation leads to the use of different regularizations, we also want to make the neural network robust to non-optimal connections which assign a noisy label to a target image.

\paragraph{Label noise.}

The cross-entropy loss has been used to get state-of-the-art classifiers on many classification tasks such as ImageNet \citep{alexnet}. However, it also comes with downsides. Using cross-entropy loss gives a huge ability to neural networks to memorize the label which leads to poor generalization properties in the case of noisy labels \citep{Zhang_2017}. In the case of domain adaptation, minibatch OT and label shift create connections between samples belonging to different classes. Going back to our example where a source minibatch is composed of one class and the target minibatch of another, \textsc{deepjdot} would transport all images from the source to the target domain even if they have different class. The loss on the classifier would be then:
\begin{equation}
    \sum_j \texttt{CE}(y_1, f(\xx_j)),\nonumber
\end{equation}
where $y_1$ is the label of images from the source minibatch. This forces the classifier to give noisy labels to the target samples and can be seen as learning with noisy labels.
To overcome this issue, we replace the cross-entropy loss 
%initially developed 
used in \citep{Damodaran_2018_ECCV} by the symmetric cross-entropy loss developed in \citep{wang2019symmetric}. This has the effect to make the classifier more robust to label noise and to non-optimal connections between source and target samples. For two probability vectors $q$ and $q^\prime$, we define the symmetric cross-entropy loss as:
\begin{align}
  %\texttt{CE}(p, q) = \sum_{i=1}^K p_i log(q_i)\\
  \texttt{SCE}(q, q^\prime) = \eta_4 \texttt{CE}(q, q^\prime) + \texttt{CE}(q^\prime, q).
\end{align}
In all our experiments we have $\eta_4=0.01$ to reduce the tuning of hyper-parameters. After describing the \textsc{deepjdot} method, MixUp regularization and the SCE loss, we have all the different ingredients to present our new methods.

\subsection{Proposed approach \textsc{MixUnbOT}}

We now present our method called \textsc{mixunbot} which is based on the methods described in the previous section. We propose to use the neighbourhing distributions $\tilde{\mu}, \tilde{\nu}$ in the OT cost from \textsc{deepjdot} like algorithms, with minibatches drawn from $\tilde{\mu}_\lambda^{\otimes m}$ and $\tilde{\nu}_\lambda^{\otimes m}$,  instead of considering the initial probability distributions $\mu$ and $\nu$. 
%, we would use their MixUp counterpart $\tilde{\mu}$ and $\tilde{\nu}$. 
By doing so, we would inherit the regularization effect of MixUp regularization on the neural network. In addition to this change, we also use the symmetric cross-entropy loss in the ground cost, instead of the standard cross-entropy loss,  to make the model robust to noisy labels from negative transfer. Thus our full loss function is:
\begin{align}
   \underset{\phi, \theta}{\operatorname{min}} &\sum_i \mathcal{L}(f_\phi(g_\theta(\xx_i^s)), \yy_i^s) + \eta_3 \mathbb{E}_{\lambda\sim \beta(\alpha), (\XX^s, \YY^s) \sim \tilde{\mu}_\lambda^{\otimes m}, \XX^t \sim \tilde{\nu}_\lambda^{\otimes m}} \quad h_{C_{\theta, \phi}}((\XX^s,\YY^s),(\XX^t,f_\phi(g_\theta(\XX^t)))), \nonumber \\
  &  \text{with }  (C_{\theta, \phi})_{i,j}=\eta_1 \|g_\theta(\xx_i^s) - g_\theta(\xx_j^t)\|_2^2 + \eta_2 \texttt{SCE}(\yy_i^s, f_\phi(g_\theta(\xx_j^t))).
\end{align}
In the case where $h$ is the exact optimal transport cost as in \textsc{deepjdot}, we call our method \textsc{mixot}. When $h$ is the unbalanced OT cost, we call our method \textsc{mixunbot}. When connections are made between samples with different labels due to OT, our proposed regularization would help the neural network not to learn the label. We show that it is the combination of the two methods which leads to an improvement in the performances in an ablation study on domain adaptation (see Section \ref{par:analysis}). Our method is fully described in Figure \ref{fig:mixot}. After presenting our method and the different related methods, we now evaluate it on different domain adaptation scenarios.

\begin{table}[t]
\small
\begin{minipage}{.5\linewidth}
    \begin{center}
        \begin{tabular}{|@{\hskip3pt}c@{\hskip3pt}|@{\hskip3pt}c@{\hskip3pt}|@{\hskip3pt}c@{\hskip3pt}|@{\hskip3pt}c@{\hskip3pt}|@{\hskip3pt}c@{\hskip3pt}|@{\hskip3pt}c@{\hskip3pt}|}
             \hline
            \multirow{8}{*}{Exp.} & Methods/DA & U $\mapsto$ M & M $\mapsto$ MM & S $\mapsto$ M & Avg\\
             
             \hline
            &\textsc{dann} & 93.2 $\pm$ 0.2 & 91.7 $\pm$ 6.5 & 89.2 $\pm$ 0.5 & 91.4 \\
            &\textsc{cdan-e} & \textbf{99.2 $\pm$ 0.1} & 96.0 $\pm$ 1.5 & 95.6 $\pm$ 1.8 & 96.9 \\
            &\textsc{alda} & 97.9 $\pm$ 1.3 & \textbf{96.7 $\pm$ 0.6} & 95.3 $\pm$ 1.5 & 96.6 \\
            & \textsc{deepjdot} & 95.9 $\pm$ 0.1 & 81.2 $\pm$ 2.0 & 94.2 $\pm$ 0.4 & 90.3\\
            & \textsc{jumbot} & 98.3 $\pm$ 0.4 & 96.6 $\pm$ 0.4 & 98.8 $\pm$ 0.2 & 97.9\\
            \cline{2-6}
             &\textsc{mixot} & 97.6 $\pm$ 0.1 & 86.2 $\pm$ 5.0 & 98.2 $\pm$ 0.1 & 94.0\\
             & \textsc{mixunbot} & 98.8 $\pm$ 0.2 & 96.4 $\pm$ 0.6 & \textbf{99.1 $\pm$ 0.1}  & \textbf{98.1}\\
             \hline
             \hline
             \multirow{4}{*}{Ablation} & \textsc{deepjdot} & 95.9 $\pm$ 0.1 & - & 94.2 $\pm$ 0.4 & 95.1\\
             & \textsc{deepjdot (rce)} & 95.9 $\pm$ 0.1 & - & 94.6 $\pm$ 0.2 & 95.3\\
             & \textsc{mixot (ce)} & 96.2 $\pm$ 0.1 & - & 94.6 $\pm$ 0.2 & 95.4\\
             & \textsc{mixot} & \textbf{97.6 $\pm$ 0.1} & - & \textbf{98.2 $\pm$ 0.1} & \textbf{97.9}\\
             \hline
        \end{tabular}
        \label{tab:digits_results}
    \end{center}
\end{minipage}
\qquad
\begin{minipage}{.45\linewidth}
    \begin{center}
        \begin{tabular}{|c|c|}
        \hline
        Methods         & Accuracy (in \%) \\ \hline
        CDAN-E     & 70.1             \\ 
        ALDA       & 70.5             \\ 
        DEEPJDOT   & 68.0             \\ 
        ROBUST OT  & 66.3             \\ 
        JUMBOT          & 70.2             \\ 
        \textsc{mixunbot} (Ours) & \textbf{71.3}             \\ \hline
        \end{tabular}
        %\caption{}
        \label{tab:VisDA_results}
    \end{center}
\end{minipage}
\caption{DA results on digit and Visda-2017 datasets. (Left table) The Above part sums up experiments on DA digits. Each experiment was run three times with different seeds and the best score is reported in \textbf{bold}. Lower parts gather an ablation study of method over two DA tasks on digits datasets. (Right table) Summary table of DA results on Visda-2017 datasets (ResNet-50). All results are reported at the end of training.}
\label{tab:digits_VisDA_results}
%\vspace{-0.1cm}
\end{table}

\section{Experiments}\label{sec:exp}
In this section, we evaluate our methods on several standard, real-world problems of domain adaptation and partial domain adaptation. For our experiments, we relied on the PyTorch deep learning library \citep{paszke2017automatic} as well as the POT library for the optimal transport solvers \citep{Flamary2021}.   
\subsection{Vanilla Domain Adaptation}
\begin{table*}[t!]
    \center
    \small
  \begin{tabular}{|@{\hskip3pt}c@{\hskip3pt}|@{\hskip3pt}c@{\hskip3pt}|c|c|c|c|c|c|c|c|c|c|c|c|c|}
    \hline
     \multirow{10}{*}{\textsc{da}} & Method & A-C & A-P & A-R & C-A & C-P & C-R & P-A & P-C & P-R & R-A & R-C & R-P & avg\\
     \hline
     & \textsc{resnet}-50 & 34.9 & 50.0 & 58.0 & 37.4 & 41.9 & 46.2 & 38.5 & 31.2 & 60.4 & 53.9 & 41.2&  59.9 & 46.1\\
     %DANN (2016) & 44.2 & 59.4 & 69.5 & 48.5 & 60.2 & 63.4 & 47.5 & 42.7 & 69.6 & 63.4 & 52.0 & 78.1 & 58.2 \\
     & \textsc{dann}  & 44.3 & 59.8 & 69.8 & 48.0 & 58.3 & 63.0 & 49.7 & 42.7 & 70.6 & 64.0 & 51.7 & 78.3 & 58.3 \\
     %JAN (2017) & 45.9 & 61.2 & 68.9 & 50.4 & 59.7 & 61.0 & 45.8 & 43.4 & 70.3 & 63.9 & 52.4 & 76.8 & 58.3\\
     & \textsc{cdan-e} & 52.5 & 71.4 & 76.1 & 59.7 & 69.9 & 71.5 & 58.7 & 50.3 & 77.5 & 70.5 & 57.9 & 83.5 & 66.6 \\
     %TAT (2019) & 51.6 & 69.5 & 75.4 & 59.4 & 69.5 & 68.6 & 59.5 & 50.5 & 76.8 & \textbf{70.9} & 56.6 & 81.6 & 65.8 \\
     %CDAN + Transnorm (2019) & 50.2 & 71.4 & 77.4 & 59.3 & 72.7 & 73.1 & 61.0 & 53.1 & 79.5 & 71.9 & 59.0 & 82.9 & 67.6\\
     %& \textsc{deepjdot} (*) & 51.4 & 69.6 & 77.0 & 61.2 & 68.9 & 71.3 & 58.2 & 48.5 &75.9 & 67.2 & 55.0 & 80.5 & 65.4\\
     & \textsc{deepjdot} & 50.7 & 68.6 & 74.4 & 59.9 & 65.8 & 68.1 & 55.2 & 46.3 & 73.8 & 66.0 & 54.9 & 78.3 & 63.5\\
     %& \textsc{e-deepjdot} & 50.6 & 68.9 & 74.4 & 59.3 & 65.1 & 69.0 & 56.2 & 46.5 & 74.5 & 65.1 & 54.7 & 78.1 & 63.5\\
     & \textsc{alda} & 52.2 & 69.3 & 76.4 & 58.7 & 68.2 & 71.1 & 57.4 & 49.6 & 76.8 & 70.6 & 57.3 & 82.5 & 65.8\\
     %DeepJDUOT ($\tau=10$) & 52.2 & 70.3 & 77.3 & 60.8 & 69.7 & 71.4 & 58.5 & 49.0 &76.2 & 68.0 & 54.8 & 80.6 & 65.7\\
     %DeepJDUOT ($\tau=5$) & 52.8 & 72.2 & 77.5 & 61.4 & 70.2 & 72.7 & 58.6 & 51.0 & 76.2 & 70.3 & 55.8 & 81.9 & 66.7\\
     %ROT(param paper) & 41.8 & 70.5 & 77.3 & 60.8 & 69.8 & 71.9 & 55.0 & 38.7 & 77.2 & 69.1 & 49.6 & 81.1 & 63.6 \\
     & \textsc{rot} & 47.2 & 71.8 & 76.4 & 58.6 & 68.1 & 70.2 & 56.5 & 45.0 & 75.8 & 69.4 & 52.1 & 80.6 & 64.3 \\
     %DeepJDUOT & \textbf{54.5} & \textbf{73.0} & \textbf{78.5} & \textbf{63.1}  & \textbf{71.9} &  \textbf{72.4}  & \textbf{60.9}  & \textbf{50.8} & \textbf{77.5} & 70.2 & 56.7 & 82.0 & \textbf{67.6}\\
     & \textsc{jumbot} & 55.2 & 75.5 & 80.8 & 65.5  & 74.4 &  74.9  & 65.2  & 52.7 & 79.2 & 73.0 & 59.9 & 83.4 & 70.0\\
     \hline
     
    % ===========================[Before]===========================
    %  & \textsc{mixot} (ours) & \textbf{56.4} & \Scd{76.8} & \Scd{81.5} & \Scd{66.4} & \textbf{76.2} & \Scd{77.0} & \Scd{64.0} & \Scd{53.5} & \Scd{81.3} & \Scd{72.9} & \Scd{58.6} & \Scd{84.4} & \Scd{70.8}\\
    %  & \textsc{mixunbot} (ours) & \Scd{55.7} & \textbf{77.6} & \textbf{82.1} & \textbf{67.1} & \Scd{75.8} & \textbf{78.1} & \textbf{66.8} & \textbf{54.5} & \textbf{81.9} & \textbf{75.1} & \textbf{60.1} & \textbf{85.4} & \textbf{71.7}\\

     % ===========================[After]===========================
    & \textsc{mixot} (ours) & \textbf{56.4} & \Scd{76.8} & \textbf{81.5} & \Scd{66.4} & \textbf{76.2} & \textbf{77.0} & \Scd{64.0} & \Scd{53.5} & \Scd{81.3} & \Scd{72.9} & \Scd{58.6} & \Scd{84.4} & \Scd{70.8}\\  
    & \textsc{mixunbot} (ours) & \Scd{56.0} & \textbf{77.5} & \Scd{81.4} & \textbf{68.4} & \Scd{74.4} & \Scd{76.7} & \textbf{66.7} & \textbf{54.2} & \textbf{81.5} & \textbf{76.3} & \textbf{60.4} & \textbf{84.9} & \textbf{71.5}\\
          
    %& \textsc{mixunbot} KF(ours) & \textbf{56.8} & \textbf{77.9} & \textbf{81.7} & \textbf{68.6} & 75.4 & \textbf{77.1} & \textbf{66.4} & \textbf{55.1} & \textbf{81.9} & \textbf{74.7} & 59.5 & \textbf{84.5} & \textbf{71.6}\\

     \hline
     & \textsc{mixot} (oracle) & 56.5 & 76.9 & 81.7 & 66.8 & 76.2 & 77.0 & 65.3 & 53.9 & 81.5 & 73.5 & 58.9 & 84.4 & 71.1\\
     & \textsc{mixunbot} (oracle) & 56.1 & 77.8 & 81.4 & 68.8 & 74.8 & 77.1 & 66.7 & 54.5 & 81.5 & 76.6 & 60.4 & 85.0 & 71.7\\

     \hline
     \hline
     \multirow{4}{*}{\textsc{pda}} & \textsc{resnet-50} & 46.3 & 67.5 & 75.9 & 59.1 & 59.9 & 62.7 & 58.2 & 41.8 & 74.9 & 67.4 & 48.2 & 74.2 & 61.4\\
     %& \textsc{deepjdot}(*) & 46.5 & 62.9 & 73.7 & 55.1 & 55.1 & 63.2 & 56.6 & 43.5 & 70.2 & 65.2 & 49.5 & 71.4 & 59.4 \\
     & \textsc{deepjdot} & 48.2 & 66.2 & 76.6 & 56.1 & 57.8 & 64.5 & 58.3 & 42.7 & 73.5 & 65.7 & 48.2 & 73.7 & 60.9 \\
     %& \textsc{e-deepjdot}(*) & 47.6 & 67.0 & 77.3 & 57.1 & 57.9 & 65.4 & 58.1 & 41.3 & 74.4 & 66.4 & 47.7 & 75.1 & 61.3 \\
     & \textsc{ba3us} & 56.7 & 76.0 & \Scd{84.8} & 73.9 & 67.8 & \Scd{83.7} & 72.7 & 56.5 & 84.9 & 77.8 & 64.5 & 83.8 & 73.6 \\
     & \textsc{jumbot} & \Scd{62.7} & \Scd{77.5} & 84.4 & \Scd{76.0} & \Scd{73.3} & 80.5 & \Scd{74.7} & \Scd{60.8} & \Scd{85.1} & \Scd{80.2} & \Scd{66.5} & \Scd{83.9} & \Scd{75.5}\\
     \hline
     & \textsc{mixunbot} (ours) & \textbf{63.3} & \textbf{80.0} & \textbf{89.0} & \textbf{78.9} & \textbf{76.6} & \textbf{87.5} & \textbf{76.4} & \textbf{65.9} & \textbf{88.3} & \textbf{81.4} & \textbf{67.6} & \textbf{86.8} & \textbf{78.5}\\
     \hline
     \multirow{10}{*}{\textsc{pda}} & \textsc{pada} & 51.9 & 67.0 & 78.7 & 52.2 & 53.8 & 59.0 & 52.6 & 43.2 & 78.8 & 73.7 & 56.6 & 77.1 & 62.1\\
     \multirow{10}{*}{oracle}& \textsc{iwan} & 53.9 & 54.4 & 78.1 & 61.3 & 47.9 & 63.3 & 54.2 & 52.0 & 81.3 & 76.5 & 56.7 & 82.9 & 63.5\\
     & \textsc{san} & 44.4 & 68.7 & 74.6 & 67.5 & 65.0 & 77.8 & 59.8 & 44.7 & 80.1 & 72.2 & 50.2 & 78.7 & 65.3\\
     & \textsc{drcn} & 54.0 & 76.4 & 83.0 & 62.1 &  64.5 & 71.0 & 70.8 & 49.8 & 80.5 & 77.5 & 59.1 & 79.9 & 69.0\\
     & \textsc{etn} & 59.2 & 77.0 & 79.5 & 62.9 & 65.7 & 75.0 & 68.3 & 55.4 & 84.4 & 75.7 & 57.7 & 84.5 & 70.4\\
     & \textsc{safn} & 59.0 & 76.2 & 81.4 & 70.4 & 73.0 & 77.8 & 72.4 & 55.3 & 80.4 & 75.8 & 60.4 & 79.9 & 71.8\\
     & \textsc{rtnet} & \Scd{63.2} & \textbf{80.1} & 80.7 & 66.7 & 69.3 & 77.2 & 71.6 & 53.9 & 84.6 & 77.4 & 57.9 & 85.5 & 72.3\\
     & \textsc{jumbot} & 62.7 & 77.8 & 86.1 & \Scd{76.1} & 73.3 & 80.5 & 72.9 & 60.8 & 85.1 & 80.2 & \Scd{66.5} & 83.9 & 75.5 \\
     & \textsc{ba3us} & 60.6 & 83.2 & 88.4 & 71.8 & 72.8 & 83.4 & 75.4 & 61.6 & 86.5 & 79.2 & 62.8 & \Scd{86.0} & 76.0\\
     & \textsc{ar} & 62.1 & 79.2 & \textbf{89.1} & 73.9 & \Scd{75.6} & \Scd{84.4} & \textbf{78.4} & \Scd{61.9} & \Scd{87.9} & \textbf{82.2} & 65.4 & 85.3 & \Scd{77.1} \\
     \hline
     & \textsc{mixunbot} (ours) & \textbf{63.3} & \Scd{80.0} & \textbf{89.1} & \textbf{78.9} & \textbf{76.7} & \textbf{87.5} & \Scd{76.4} & \textbf{65.9} & \textbf{88.3} & \Scd{81.5} & \textbf{67.8} & \textbf{86.8} & \textbf{78.6}\\
     %& \textsc{mixunbot} (2)& \textbf{63.4} & \textbf{82.1} & \textbf{90.0} & \textbf{76.1} & \textbf{73.6} & \textbf{86.6} & \textbf{75.5} & \textbf{63.1} & \textbf{89.2} & \textbf{82.7} & \textbf{67.7} & \textbf{87.3} & \textbf{78.1}\\
     \hline
\end{tabular}
\vspace{-0.2cm}
  \caption{(Best view in color) Summary table of domain adaptation and partial domain adaptation results on Office-Home dataset with a ResNet-50. Results are reported at the end of training except for PDA oracle, where best score along training is reported. \textbf{Best scores} are in \textbf{bold} and \Scd{second best scores} are in \Scd{blue}.}
  \label{tab:office_home}
    \vspace{-0.3cm}
\end{table*}

In this paragraph, we evaluate and discuss the practical advantage of our proposed methods on vanilla domain adaptation experiments. In vanilla domain adaptation, the source and target domains share the same classes with a likely shift in the labels.

{\bfseries Datasets.} We start the experimental section with \textbf{digits} datasets. We follow the evaluation protocol of \citep{Damodaran_2018_ECCV} on three domain adaptation scenarios based on digits datasets: USPS
to MNIST (U$\mapsto$M), MNIST to MNIST-M (M$\mapsto$MM), and SVHN
to MNIST (S$\mapsto$M). MNIST \citep{MNIST} has 60,000
images of handwritten digits, MNIST-M  has the 60,000 MNIST
images with color patches \citep{DANN} and USPS \citep{USPS} has
7,291 digit images. Street View House Numbers (SVHN)\citep{svhn} is a dataset of 73, 257 images with digits and numbers in natural scenes. We report the evaluation results on the test target datasets at the end of training.
\textbf{Office-Home} \citep{office_home} is a real world dataset for unsupervised domain adaptation (UDA) composed of 15,500 images from four different domains: Artistic
(A), Clip Art (C), Product (P) and Real-World
(R) images. For each domain, the dataset has images of 65
object categories that are common in office and home scenarios. We evaluate our methods in the 12 possible scenarios.
Our last dataset is \textbf{VisDA-2017} \citep{visda}. It is a large-scale dataset from simulation to real images. Visda-2017 contains 152,397 synthetic images as the source domain and 55,388 real-world images as the target domain. There are 12 different object categories in the
two domains. Following \citep{cdan2018, alda2020}, we evaluate all methods on Visda-2017 validation set.

\paragraph{Results.} We compare our \textsc{mixot} methods against state-of-the-art methods: \textsc{dann}\citep{DANN}, \textsc{cdan-e} \citep{cdan2018}, \textsc{alda} \citep{alda2020}, \textsc{rot} \citep{balaji2020robust}. We also consider the methods which are at the heart of \textsc{mixot} variants, namely \textsc{deepjdot} \citep{Damodaran_2018_ECCV} and \textsc{jumbot} \citep{fatras21a}. We reported the accuracy on the testset at the end of training for all benchmarks and methods. 
We believe it is a more fair method than reporting the best results along training on the testset, because evaluating the method along training breaks the assumptions on labels on the target domain. However, on Office-Home, we reported the results with the highest prediction along training for our method to show that we are not prone to overfitting (see when oracle is mentioned). As our setting is identical to \cite{fatras21a}, we reported their results for the different competitors on Office-Home and VisDA-2017. The full details of our training procedure and architectures can be found in appendix.

The results on digit datasets can be found in the left above part of Table \ref{tab:digits_VisDA_results}. We conducted each experiment three
times, with different seeds, and report their average results and variance. We only resize and
normalize the image without data augmentation. We see that our method \textsc{mixot} performs $4\%$ better than \textsc{deepjdot} on average while \textsc{mixunbot} performs slightly better than \textsc{jumbot}. Furthermore, we see an important $8\%$ increase of the performance compared to \textsc{deepjdot}. Regarding Office-Home experiments, the results are gathered in the above part of Table \ref{tab:office_home}. For fair comparison with previous
work, we used a similar data pre-processing, training procedure, architectures, hyperparameter selection on the AC task, and we used the ten-crop technique \citep{cdan2018, alda2020} for testing our methods. Our method \textsc{mixunbot} achieves better performances on all considered tasks than all competitors and is on average 2$\%$ higher than the best competitor \textsc{jumbot}. It achieves the best performances on 10 tasks over 12 and is only beaten by our method \textsc{mixot}. Regarding \textsc{mixot}, it achieves a better accuracy than \textsc{jumbot} on average and on 10 of the 12 scenarios from the Office-Home dataset, furthermore, it is higher than \textsc{deepjdot} by more than $7\%$ showing the regularization effect of our strategy. When \textsc{mixot} does not achieve the best score, it is in second position behind \textsc{mixunbot}. Finally, on Visda-2017 datasets, the results can be found in the right part of Table \ref{tab:digits_VisDA_results}. \textsc{mixunbot} performs better than all competitors, in particular, it is 1$\%$ above our reproduced \textsc{jumbot} score. 

\paragraph{Analysis and sensitivity study.}\label{par:analysis} %TO DO (TSNE, classification accuracy along training, ablation)
In this paragraph, we provide an empirical analysis of our method. We conducted an ablation study between \textsc{deepjdot} and \textsc{mixot} as the difference between \textsc{mixunbot} and  \textsc{mixot} is the use of unbalanced OT instead of exact OT at the minibatch level. The ablation study can be found in the lower left part of Table \ref{tab:digits_VisDA_results}. The difference between \textsc{deepjdot} and \textsc{mixot} is first the use of the symmetric cross-entropy loss instead of cross-entropy loss between labels in the transfer loss. The second difference is the use of MixUp data augmentation on source and target samples. As we can see in the results, the use of the SCE loss in \textsc{deepjdot} and the use of MixUp data with the standard cross-entropy loss do not improve results over \textsc{deepjdot}. However, when the MixUp regularization is coupled with the SCE loss, we see a 1.7$\%$ increase on the \textsc{usps} to \textsc{mnist} task and a 3.5$\%$ increase on the \textsc{svhn} to \textsc{mnist} task, demonstrating the practical advantage of our approach. The ablation study shows that we can not take fully advantage of the MixUp regularization with noisy labels. We conjecture that the success of the combination is due to the SCE loss which prevents the neural network to overfit on noisy labels. It allows MixUp to regularize the neural network as it would have done in a clean label environment.

\begin{figure}[t]
    \centering
        \includegraphics[width=1.\linewidth]{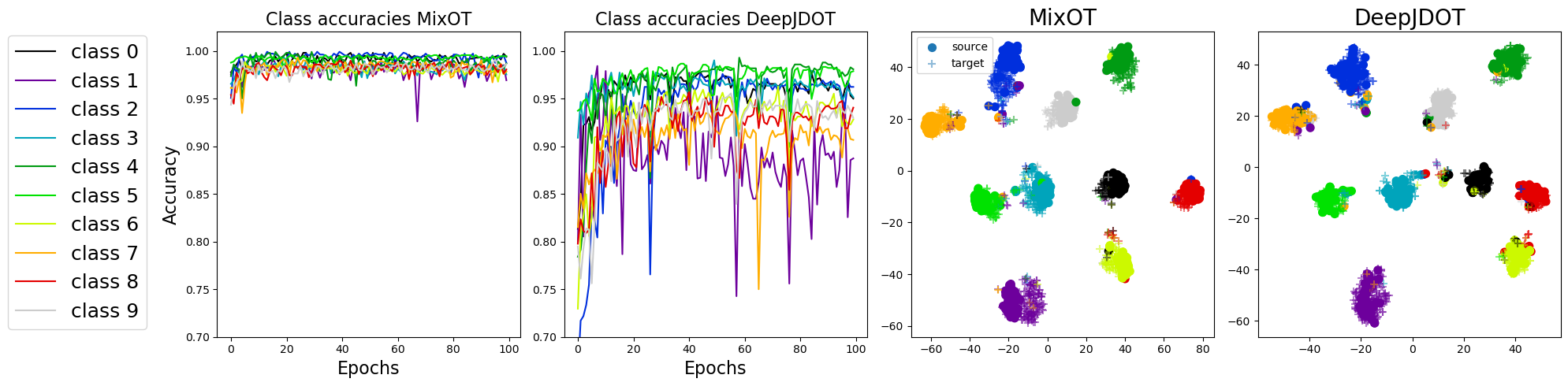}
    \vspace{-0.5cm}
    \caption{(Best view in color) (Two left plots) Classification accuracy on classes along training for the SVHN to MNIST DA task of our method and \textsc{deepjdot}. The performances of our method reach higher accuracies than \textsc{deepjdot} and are more stable. (Two right plots) T-SNE embeddings of 1500 test samples from SVHN (source) and MNIST (target) for \textsc{deepjdot} and our method \textsc{mixot} at test time. It shows the ability of the methods to discriminate classes (samples are colored w.r.t. their classes). }
    \label{fig:tsne}
    \vspace{-0.4cm}
\end{figure}

We also provide a classification accuracy along training as well as a T-SNE plot \citep{vandermaaten08a} on embedded source and target test samples between \textsc{deepjdot} and \textsc{mixot} (see Figure \ref{fig:tsne}). Regarding the classification accuracy along training, we see that our method reach higher accuracies on classes and that these acuracies are stable with a low variance. However for \textsc{deepjdot}, we see that the training is not stable, with some variance and a small overfitting issue on some classes after 50 epochs. Moreover, the accuracy on some classes is lower than the ones achieved by our method as justified in our experiments. Regarding the TSNE plot, we can see that our method produces well defined clusters while the \textsc{deepjdot} method has some overlaps between them. 

\begin{figure}[t]
    \centering
        \includegraphics[width=1.\linewidth]{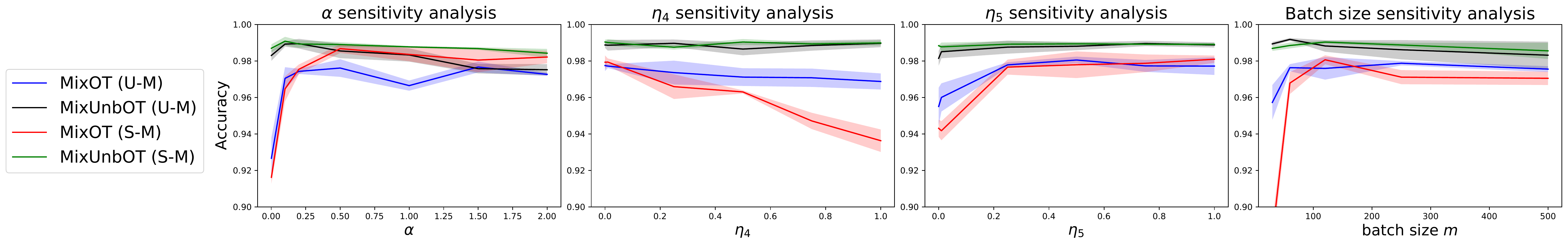}
        \vspace{-0.3cm}
    \caption{(Best view in color) Sensitivity analysis of our methods \textsc{mixot} and \textsc{mixunbot} for several hyperparameter variations ($\alpha$, $\eta_4, \eta_5$ and batch size $m$). We report the average classification accuracy and variance on three runs with different seeds for DA tasks \textsc{usps} $\mapsto$ \textsc{mnist} (U-M) and  \textsc{svhn} $\mapsto$ \textsc{mnist} (S-M). Networks have been pretrained for 2 epochs except on (U-M) batch size sensitivity analysis (10 epochs).}
    \label{fig:sensitivity}
    \vspace{-0.3cm}
\end{figure}

Finally, we conducted a sensitivity analysis on the DA tasks \textsc{USPS} to \textsc{MNIST} and \textsc{SVHN} to \textsc{MNIST} for the following hyper-parameters: coefficients $\alpha$, $\eta_4, \eta_5$ and batch size $m$. $\eta_4, \eta_5$ are constants in the SCE loss $\texttt{SCE}(q, q^\prime) = \eta_4\texttt{CE}(q, q^\prime) + \eta_5\texttt{CE}(q^\prime, q)$ to evaluate the influence of the term $\texttt{CE}(q^\prime, q)$. All results are summarized in Figure \ref{fig:sensitivity}. We have higher accuracies for values of $\alpha$ greater than 0.1 and smaller than 0.5 as larger values lead to high perturbed samples. Regarding $\eta_4$, we can see that the accuracy remains stable for \textsc{mixunbot} and decreases for \textsc{mixot} when $\eta_4$ gets closer to 1. We also see a decrease in the accuracy for small values $\eta_5$, \emph{i.e.,} when we recover the typical cross-entropy. It shows that MixUp leads to improvement when used with the SCE loss. Finally, we can see that small minibatch sizes lead to poor performances for \textsc{mixot} probably due to the high negative transfer between domains.% and we remark that the performances of \textsc{mixunbot} are low for large batch size, probably due to a non meaningful pre-trained network.

%\subsubsection*{Author Contributions}
%If you'd like to, you may include  a section for author contributions as is done
%in many journals. This is optional and at the discretion of the authors.

%\textbf{This should be only done in the camera ready version of the manuscript, not in the anonymized version submitted for review!}

%\subsubsection*{Acknowledgments}
%Use unnumbered third level headings for the acknowledgments. All
%acknowledgments, including those to funding agencies, go at the end of the paper.

%\textbf{This should be only done in the camera ready version of the manuscript, not in the anonymized version submitted for review!}

\subsection{Partial Domain Adaptation}

In this section, we study the performances of our proposed method \textsc{mixunbot} on partial domain adaptation (PDA) experiments. In partial domain adaptation, some classes in the source domain do not exist in the target domain and thus, the target labels are a subset of the source labels, \emph{i.e.,
$\mathcal{Y}_t \subset \mathcal{Y}_s$}. The different samples which are in these classes can be considered as outliers at the level of the target domain as they would produce negative transfer. That is why methods based on exact Optimal Transport fail to solve this problem as they would transport these samples to the target domain, however \textsc{jumbot} has been shown to perform well on this problem thanks to Unbalanced OT. We thus want to evaluate the performances of \textsc{mixunbot} in partial DA. 

We evaluate our method on the dataset partial Office-Home. We follow the setting from \cite{Cao_2018_ECCV} which
selects the first 25 categories (in alphabetic order) as target classes to form the partial target
domain. We compare our method against the state-of-the-art partial domain adaptation methods:
\textsc{pada} \citep{Cao_2018_ECCV}, \textsc{etn} \citep{ETN_2019_CVPR}, 
\textsc{ba3us} \citep{liang2020baus}, \textsc{safn} \citep{Xu_2019_ICCV}, \textsc{rtnet} \citep{Chen2020SelectiveTW}, \textsc{iwan} \citep{Zhang2018ImportanceWA}, \textsc{san} \citep{Cao2018PartialTL}, \textsc{drcn} \citep{drcn2021}, \textsc{jumbot} \citep{fatras21a} and \textsc{ar} \citep{gu2021adversarial}. For fair comparison we followed the
experimental setting of \textsc{pada}, \textsc{etn}, \textsc{ba3us} and \textsc{jumbot}. We selected the best hyper parameters on the A-C task as validation and used them on the remaining tasks. We
report all the training and architecture details to the appendix. We evaluate our methods in two different ways. We first report the performances of our method at the end of training against  \textsc{ba3us}, \textsc{deepjdot} and \textsc{jumbot}. These results are reported from \citep{fatras21a}. Then we report the best score along training of our methods (oracle), where we evaluated the models every 1250 iterations, against all mentioned competitors where we reported the scores from their paper for our setting (neural network architecture and training procedure). All results can be found in the lower part of Table \ref{tab:office_home}.

For results at the end of training, we can see that \textsc{mixunbot} achieves the highest performances on all partial domain adaptation tasks and is on average 3$\%$ above the best competitor \textsc{jumbot}, showing the relevance to regularize the neural networks as we do. Regarding the oracle score, we see that \textsc{mixunbot} achieves the highest performances on 10 partial domain adaptation tasks over 12 and that our method is $1.5\%$ above the best competitor on average. The different results between our oracle score and the score at the end of training are similar on each task which show that our method does not suffer from an overfitting issue in the training.

\section{Conclusion}

In this paper we proposed new methods to tackle the domain adaptation problem. When label shift occurs between domains,  optimal transport based methods create negative alignments. These negative alignments assign noisy labels to target samples which can decrease the performances of models. We proposed a new method to mitigate the influence of these noisy labels on our models. We first used the MixUp regularization on both source and target domains. We then used the symmetric cross-entropy loss, a noisy label robust loss function, when aligning the source and the target distributions. Afterwards, we evaluated our methods on several benchmark and real-world domain adaptation and partial domain adaptation scenarios. We showed that our methods are able to outperform recent state-of-the-art methods. In an ablation study, we showed that it is the combination of the MixUp regularization and the symmetric cross-entropy loss which increases performance, while individually they do not improve results. Future work will consider a mathematical analysis on the use of MixUp augmentation and symmetric cross-entropy to justify the empirical performances or the use of manifold MixUp regularization \citep{verma19a}. 

\subsubsection*{Author Contributions}
In this section, we state the contributions of each author on each part of the present manuscript.
\begin{itemize}
    \item Formalism : KF
    \item Proof-of-concepts : KF
    \item Experiments : KF, HN
    \item Experiments review : KF, HN
    \item Writing--original draft preparation : KF
    \item Writing—review and editing : KF, HN, IM
    \item Supervision : KF, IM
    \item Funding acquisition : IM
\end{itemize}

\subsubsection*{Acknowledgments}
This work was partially supported by NSERC Discovery grant (RGPIN-2019-06512), a Samsung grant and a Canada CIFAR AI chair. Authors thank Nicolas Courty and R\'emi Flamary for fruitful discussions.

%Due to the marginal constraints, optimal transport can transport samples from different classes, especially when there is a shift in the labels between domains. Furthermore its minibatch approximation, which is widely used in practice, is even more sensitive to this issue. These connections create negative alignments between domains and deteriorates the performances of neural networks. 

%\section{Societal impact}
%The contribution of this paper is to bring knowledge on how one can regularize neural networks to mitigate the effect of negative transfer on domain adaptation problems. As in domain adaptation one wants to predict the label of target samples when no label is available, efficient methods can potentially lead to privacy issues. 
\bibliography{collas2022_conference}
\bibliographystyle{collas2022_conference}

\newpage
\appendix
\section*{Appendix}

\section{Additional experiments}
\begin{figure}[t]
    \centering
        \includegraphics[width=1.\linewidth]{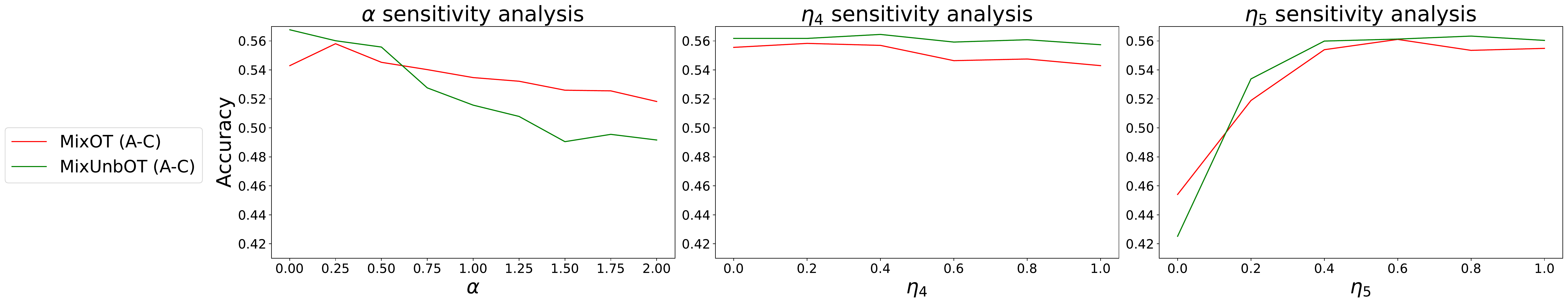}
        \vspace{-0.3cm}
    \caption{(Best view in color) Sensitivity analysis of our methods \textsc{mixot} and \textsc{mixunbot} for several hyperparameter variations ($\lambda$, $\eta_4, \eta_5$). We report the classification accuracy on the AC task of the Office-Home dataset.}
    \label{fig:sensitivity_AC}
    \vspace{-0.3cm}
\end{figure}

In this section, we include a sensitivity analysis of our methods on the AC task of the Office-Home dataset (see Figure \ref{fig:sensitivity_AC}).

\section{Experimental Settings}
\label{appendix:setting}

\subsection{Implementation and Environment for Experiment}
% We perform our experiment with Graham Cluster \footnote{\url{https://docs.computecanada.ca/wiki/Graham}}. 

Experiments were conducted on two 1080-Ti NVIDIA GPUs. As a software environment, we use CentOS 7, gcc 9.3.0, Python 3.8.2, Pytorch 1.10.0, cuDNN 8.2.0, and CUDA 11.4. 
Our code can be found here \url{https://github.com/kilianFatras/MixOT}.
% Our code can be found at the link below.\\
% \url{https://github.com/kilianFatras/MixOT} \hiroki{(coming soon. or anonymize?)}

\subsection{Symmetric cross-entropy loss}
For two probability vectors $q, q^\prime \in \Sigma_k$, where $\Sigma_k$ is the simplex of dimension $k$, the symmetric cross-entropy loss is defined as:
\begin{align}
  \texttt{SCE}(q, q^\prime) = \texttt{CE}(q, q^\prime) + \texttt{CE}(q^\prime, q),\nonumber\\
  \text{ where } \texttt{CE}(q, q^\prime) = \sum_{i=1}^K q_i log(q_i^\prime). \nonumber
\end{align}

However when $q$ is a one-hot vector, like a label for instance, $\texttt{CE}(q^\prime, q)$ is not directly computable because of the logarithm. That's why we clip the value of $q$ with a small constant of $1e^{-7}$ in our experiments. A similar strategy was used in \cite{wang2019symmetric}.

\subsection{Unbalanced OT}

Our method \textsc{mixunbot} is based on unbalanced optimal transport. In practice, we used the entropically-regularized variant of Unbalanced OT as done in \cite{fatras21a}. It allows us to use a generalized formulation of the Sinkhorn algorithm \citep{CuturiSinkhorn, ChizatPSV18} which is simple to implement in practice. For digits experiments the entropic regularization coefficient was set to 0.1. For the VisDA and Office Home experiments, it was set to 0.01. 

\subsection{Model and datasets}
In the digits experiment, we used a 6 CNN layer neural network as feature exctrator and 1 dense layer for classification as proposed in \cite{Damodaran_2018_ECCV}. The number of input channels is set to 3 for the experiments for SVHN to MNIST and MNIST to MNISTM, while the number of channels is set to 1 only for the experiments for USPS to MNIST.
In the Office-Home and Visda-2017 experiments, we use the pre-trained model of ImageNet-1K on Resnet-50 \footnote{You can find the pre-trained model at \url{https://download.pytorch.org/models/resnet50-0676ba61.pth}}.

\subsection{hyper-parameters and detailed configuration}
\label{appendix:hyper-parameters}
Note that for all experiments we relied on a stratified source minibatches, meaning that we had the same number of samples per class in source minibatches, as it was done in \citep{Damodaran_2018_ECCV,fatras21a}. 

\subsubsection{Digits experiments}
In the digits experiment, we basically followed the experimental setup from \textsc{deepjdot} \citep{Damodaran_2018_ECCV} which uses Adam as optimizer with a learning rate set to $2e^{-4}$ and without applying weight decay. 
We pre-train our neural network on the source domain for 2 epochs and then, we train our model for 100 epochs. The batch size is set to 120. The same experimental procedure is used for the ablation study the sensitivity analysis. The best hyper-parameters for \textsc{mixot} are $\eta_1=0.1, \eta_2=0.1$ and for \textsc{mixunbot} $\eta_1=0.1, \eta_2=0.1, \eta_3=1., \tau=1.$.

Regarding competitors, we use the official implementations with the considered architecture and training procedure except that we use Adam with weight decay (set to $1e^{-5}$) and a learning rate of $2e^{-4}$, since we observe that simply adaptation of Adam significantly degrades the performance of these methods, and their experiments in their paper are reported by using Momentum SGD and weight decay. The USPS to MNIST task served as validation of hyper-parameters.
%Although DANN \citep{DANN} uses the Adam optimizer without weight decay, we also experimented by using Momentum SGD with weight decay to compare with CDAN-E and ALDA.
%The sensitivities to learning rate, optimizer selection , and weight decay are compared in Section \ref{appendix:hp-sensitivity}.
%In the CDAN-E, DANN, and ALDA experiments, hyper-parameters were searched in the range of Table \ref{table:digits-hyperparams}.

%\begin{table}[h]
%\begin{center}
%\begin{tabular}{c|ccc}\hline
%     Method    & optimizer   &weight decay         & learning rate  \\\hline
%DANN    &         \{Momentum SGD, Adam\}    &    \{0, 1e-5\}    &   \begin{tabular}{c}   \{0.0001 0.00025, 0.0005, 0.00075, 0.001, 0.0025, \\0.005, 0.0075, 0.01, 0.025, 0.05, 0.075, 0.1\}    \end{tabular}　 \\

%CDAN-E    &         \{Momentum SGD, Adam\}    &    \{0, 1e-5\}    &   \begin{tabular}{c}   \{0.0001 0.00025, 0.0005, 0.00075, 0.001, 0.0025, \\0.005, 0.0075, 0.01, 0.025, 0.05, 0.075, 0.1\}    \end{tabular}　 \\

%ALDA    &         \{Momentum SGD, Adam\}    &    \{0, 1e-5\}    &   \begin{tabular}{c}   \{0.0001 0.00025, 0.0005, 0.00075, 0.001, 0.0025, \\0.005, 0.0075, 0.01, 0.025, 0.05, 0.075, 0.1\}    \end{tabular}　 \\
%\hline
%\end{tabular}
%\end{center}

%\caption{Hyperparameter Search Range: Digits Experiments for CDAN-E, DANN, and ALDA methods}
%\label{table:digits-hyperparams}

%\end{table}

\subsubsection{Office-Home experiments}

In the Office-Home experiment, the number of steps was fixed at 10K, and the batch size was 65, the same number as the number of classes as done in the \cite{fatras21a} experiments. Nesterov's acceleration method was used as an optimization method, and the momentum coefficient was 0.9. As a regularization, the weight decay is set to $5e^{-4}$, and learning rate decay is applied. We schedule the learning rate with the strategy in \cite{DANN}, it is adjusted by $\chi_p = \frac{\chi_0}{(1+\zeta l)^\kappa}$, where $l$ is the training progress linearly changing from 0 to 1, $\chi_0 = 0.01$, $\zeta = 10$, $\kappa = 0.75$. 

Since Office-Home has four domains, we used the AC task as validation to determine the best hyper-parameters. Evaluation was done using the test 10 crop technique as in \cite{fatras21a, liang2020baus}. The hyper-parameters search of the experiments shown in Table \ref{tab:office_home} of Section \ref{sec:exp} are explained in Table \ref{table:Office-Home-hyperparams}. 
Hyperparameter $\alpha$ for Beta distribution used in MixUp is set to 0.2.
The best hyper-parameters for \textsc{mixot} on vanilla Office-Home are $\eta_1=0.0025, \eta_2=0.05$.
and for \textsc{mixunbot}, the best hyper-parameters are $\eta_1=0.005, \eta_2=0.1, \tau=0.5$.

\begin{table}[h]
\begin{center}
\begin{tabular}{c|ccc}\hline
     Method    & $\eta_{1}$              & $\eta_{2}$         & $\tau$           \\\hline
\textsc{mixot}    &  \{0.0005, 0.001, 0.0025, 0.005, 0.01, 0.05, 0.1\} & \{1.0, 0.5, 0.1, 0.05, 0.01\}  &       -          \\
\textsc{mixunbot} & \{0.0001, 0.0005, 0.001, 0.005,  0.01, 0.05, 0.1\} & \{1.0, 0.5, 0.1, 0.05, 0.01\} & \{0.3, 0.5, 0.7, 1.0\}   \\\hline
\end{tabular}
\end{center}
\caption{Hyperparameter Search Range: Office-Home Dataset}
\label{table:Office-Home-hyperparams}
\end{table}

\subsubsection{Visda-2017 experiments}

Visda-2017 dataset is composed of two domains: synthetic images and real images. The Visda-2017 experiment has almost the same configuration as the Office-Home experiment. The difference is that we set the batch size to 72, which is 6 times the number of Visda-2017 classes 12.
%Our hyper-parameters are chosen by using the validation dataset. 
They are set to $\eta_1 = 0.001, \eta_2 = 0.5, \tau = 0.25$.
The hyper-parameters search range of the Visda-2017 experiments shown in Table \ref{tab:digits_VisDA_results} of Section \ref{sec:exp} are explained in Table \ref{table:visda-hyperparams}.

\begin{table}[h]
\begin{center}
\begin{tabular}{c|ccc}\hline
     Method    & $\eta_{1}$              & $\eta_{2}$         & $\tau$           \\\hline
% \textsc{mixot}    &  \{0.0001, 0.005, 0.001, 0.05, 0.1\} & \{0.005, 0.01, 0.05, 0.1, 0.5\}  &       -                 \\
\textsc{mixunbot} & \{0.0001, 0.005, 0.001, 0.05, 0.1\} & \{0.005, 0.01, 0.05, 0.1, 0.5\} & \{0.25, 0.5, 0.7\}  \\\hline
\end{tabular}
\end{center}
\caption{Hyperparameter Search Range: Visda-2017 Dataset}
\label{table:visda-hyperparams}
\end{table}

\subsubsection{Partial Ofice-Home}

For Partial Domain Adaptation, we considered the same neural network architecture and the same training procedure as in \cite{fatras21a, liang2020baus, Cao_2018_ECCV, ETN_2019_CVPR}. We used a pre-trained Resnet-50 on imagenet as pre-trained neural network. Our hyper-parameters are chosen by using the AC task as validation. They are set to $\eta_1 = 0.001, \eta_2 = 0.05, \tau = 0.05, $ and finally $\eta_3$ was set to 15. We trained the network for 5000 iterations with a stratified batch size of 65 (one element of each class per batch) and for optimization procedure, we used the same as in \cite{liang2020baus}. We do not use the ten crop technique to evaluate our model on the test set like \cite{fatras21a, liang2020baus}. Furthermore, we do not know if the official reported results of competitors were evaluated at the end of training or during training.

\end{document}

%% file: math_commands.tex
\usepackage{amsmath,amsfonts,bm}

\newtheorem{prop}{Proposition}

\newcommand{\yy}{{\boldsymbol{y}}}
\newcommand{\xx}{{\boldsymbol{x}}}
\newcommand{\zz}{{\boldsymbol{z}}}
\newcommand{\YY}{{\boldsymbol{Y}}}
\newcommand{\XX}{{\boldsymbol{X}}}

\newcommand{\PP}{{\boldsymbol{P}}}

\def\eqref#1{equation~\ref{#1}}
\def\1{\bm{1}}

\DeclareMathAlphabet{\mathsfit}{\encodingdefault}{\sfdefault}{m}{sl}
\SetMathAlphabet{\mathsfit}{bold}{\encodingdefault}{\sfdefault}{bx}{n}

\newcommand{\R}{\mathbb{R}}

\newcommand{\KL}{D_{\mathrm{KL}}}